\documentclass{bmvc2k}

\title{Multi-Stream Attention Learning for Monocular Vehicle Velocity and Inter-Vehicle Distance Estimation}

\addauthor{Kuan-Chih Huang}{s928001810@gmail.com}{1}
\addauthor{Yu-Kai Huang}{r08922053@ntu.edu.tw}{1}
\addauthor{Winston H. Hsu}{whsu@ntu.edu.tw}{1}

\addinstitution{
 National Taiwan University,\\
 Taipei, Taiwan
}

\runninghead{Huang ET AL}{Multi-Stream Monocular Vehicle Velocity Estimation}

\def\eg{\emph{e.g}\bmvaOneDot}

\def\etal{\emph{et al}\bmvaOneDot}

\usepackage{floatrow}
\usepackage{amssymb}
\usepackage{amsmath}

\begin{document}

\maketitle

\begin{abstract}
Vehicle velocity and inter-vehicle distance estimation are essential for ADAS (Advanced driver-assistance systems) and autonomous vehicles. To save the cost of expensive ranging sensors, recent studies focus on using a low-cost monocular camera to perceive the environment around the vehicle in a data-driven fashion. Existing approaches treat each vehicle independently for perception and cause inconsistent estimation. 
Furthermore, important information like context and spatial relation in 2D object detection is often neglected in the velocity estimation pipeline.
In this paper, we explore the relationship between vehicles of the same frame with a global-relative-constraint (GLC) loss to encourage consistent estimation. A novel multi-stream attention network (MSANet) is proposed to extract different aspects of features, e.g., spatial and contextual features, for joint vehicle velocity and inter-vehicle distance estimation.
Experiments show the effectiveness and robustness of our proposed approach. MSANet outperforms state-of-the-art algorithms on both the KITTI dataset and TuSimple velocity dataset.
\end{abstract}

\section{Introduction}
\begin{figure*}[t]
\vspace{0.3cm}
\begin{center}
\includegraphics[scale=0.68]{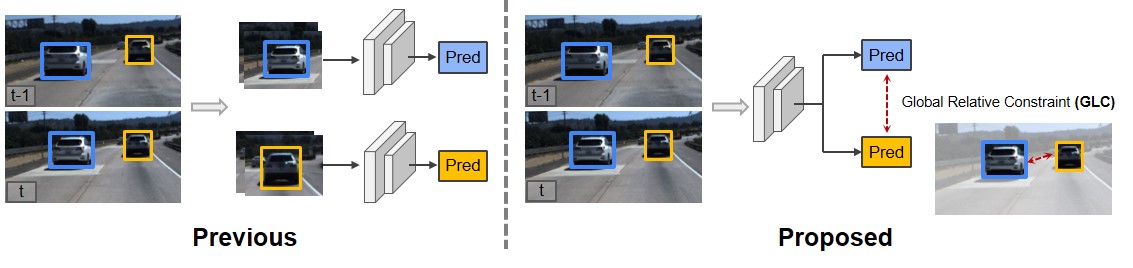}
\vspace{-1.2 cm}
\end{center}
   \caption{\textbf{Left}: Previous methods \cite{Kampelmuhler2018camera, song2020endtoend} treat each vehicle independently and estimate their velocity and position separately. \textbf{Right}: Our proposed method jointly learns each vehicle's state with the relative constraint (Section \ref{sec:loss}), which helps the network to learn the global consistency of the predictions.}
\label{fig:example}
\end{figure*}

Self-driving cars and ADAS (Advanced driver-assistance systems) have significant impacts on today’s society. Inter-vehicle distance estimation and the relative velocity between vehicles are two fundamental requirements for ADAS or autonomous vehicles to prevent the ego-vehicle from collisions. They are also crucial for path prediction, path planning, and decision making \cite{Sadat2020p3}.

Most of the 3D perception applications for the vehicles rely on ranging sensors (\eg, LiDAR, Radar), which can directly perceive the surrounding environment by emitting radio waves or laser pulse. However, these sensors have the disadvantage of sparse output, and high-cost \cite{li2019stereoRCNN}. On the contrary, a monocular camera is another good candidate for its lower affordable price. Additionally, the camera can provide dense color image information, richer texture, and a high frame rate, which is suitable for the ADAS system. Furthermore, considering the success of deep neural networks in the vehicle environment perception tasks, 
including object detection~\cite{wang2019pseudo, ma2020patchnet} and depth estimation~\cite{Eigen2014depth, Fu2018dorn}, 
exploring a new deep learning method to estimate the distance and relative velocity of the surrounding vehicle using a monocular camera is desired~\cite{song2020endtoend}.

The existing vehicle distance estimation methods based on the monocular camera can be roughly categorized into two branches: One is monocular depth estimation \cite{Eigen2014depth, Fu2018dorn}, the other is 3D monocular object detection \cite{Mousavian3Dbbox, li2020RTM3D, ma2020patchnet}. The former learns the depth of each pixel of the image by the supervised dense depth map. The latter regresses the pose of the vehicle in the world coordinate. 
Instead of predicting all the 3D bounding boxes or per-pixel depth maps, we aim at finding the nearest distance and velocity of the vehicles nearby for the purpose of ADAS application, similar to \cite{song2020endtoend}. 
This significantly reduces the effort of annotation cost and also benefits for use in real scenarios.

The vehicle velocity estimation task is widely used in traffic surveillance \cite{Jung2017resbase, Tran2018traffic}. A stationary camera is applied to estimate the velocity of the vehicle and analyze the traffic flow. However, due to the surveillance camera being fixed, the problem is less complex than the case of estimating relative velocity on the ego-vehicle. Previous works \cite{Kampelmuhler2018camera, song2020endtoend} leverage the multiple features, including depth map, optical flow, tracking information, or geometry cues, to estimate the vehicle's velocity. Though the above two works achieve remarkable results, they treat each vehicle velocity estimation problem independently and estimate each vehicle state separately, which will cause prediction inconsistency.

To address the aforementioned problem, we incorporate relative constraints and explore the relationship between each vehicle in the same frame and propose a \textit{global relative constraint (GLC) loss} shown in Figure~\ref{fig:example}. 
Instead of treating each vehicle independently, GLC loss is designed to regularize the difference between the estimation of a vehicle's relative state and the corresponding ground truth, thus benefiting learning consistent and reasonable prediction (detailed in Section \ref{sec:loss}).
Besides, observing the importance of contextual information and spatial position \cite{Choi2020hanet} for enriching the representation of the vehicle, we leverage multiple information, including context-aware features, motion clues, and spatial patterns, to jointly predict the vehicle's state. 

In summary, we make the following contributions: (1.) We propose to leverage motion, contextual, and spatial clues to extract helpful information for joint relative velocity and inter-vehicle distance estimation. 
(2.) A global relative constraint (GLC) loss $L_{rel}$ is presented to encourage the model to learn the consistency features between each vehicle in the same frame. 
(3.) Experimental results show that our approach outperforms the state-of-the-art algorithms on two public datasets.
\section{Related Work}
\label{sec:rw}
\label{sec:relatedwork}
\noindent{\bf{Monocular inter-vehicle distance estimation.}} 
To estimate the distance of the inter-vehicle, the naive way is monocular depth estimation. Several supervised learning-based methods were proposed to deal with the depth prediction tasks. Eigen \etal \cite{Eigen2014depth} presents a multi-scale network to refine the predicted depth map iteratively from different stages of the network. DORN \cite{Fu2018dorn} introduces an ordinary regression loss for depth network learning. To save the label efforts, Bian \etal \cite{bian2019depth} presents an unsupervised framework to predict the depth and ego-motion jointly in the image sequence, which significantly decreases the annotation costs.
On the other hand, some studies investigate regressing the 3D bounding box of vehicles to measure the vehicle's position. Based on the given vehicle shape prior and geometry constraint, 3DBBox \cite{Mousavian3Dbbox} estimates the vehicle's pose from a single RGB image. Pseudo-LiDAR \cite{wang2019pseudo} converts the estimated depth map to the 3D point cloud, which can be utilized to predict 3D bounding boxes by existing 3D detectors. PatchNet \cite{ma2020patchnet} figures out that projecting 2D bounding box into world coordinate can help to learn the 3D vehicle position. 

\smallskip\noindent{\bf Monocular vehicle velocity estimation.}
To our best knowledge, there are few works developing monocular velocity estimation for the vehicle. In \cite{Tran2018traffic}, a fixed camera is utilized to capture the traffic flow. Based on the geometry constraint and the known location of the landmarks, the velocity of the vehicle can be calculated by the parameters of the camera calibration, which is simpler than the application on the onboard vehicle. 
To estimation the relative velocity from ego-vehicle, Kampelmühler \etal \cite{Kampelmuhler2018camera} directly regress the vehicle velocity from monocular sequences, which utilize several cues like motion feature from Flownet \cite{Ilg2017flownet2} and depth feature from Monodepth \cite{Godard2017monodepth}. 
Furthermore, Song \etal \cite{song2020endtoend} utilize geometry constraint and optical flow feature to predict the velocity and distance of vehicle jointly.
Though the above works obtain the desired performance, they predict each vehicle's state individually (Figure \ref{fig:example}), which neglects to explore the relationships between neighboring vehicles. Furthermore, they do not explore additional cues like context-aware or spatial information. 

\smallskip\noindent{\bf Self-attention mechanism.}
Our multi-stream fusion block is related to the self-attention mechanism \cite{vaswani2017SA}.
The self-attention mechanism has been widely leveraged in sequential modeling and has considerable improvement in natural language processing (NLP) tasks. Compared with CNN, RNN, and LSTM, self-attention can achieve better performance due to its capability to capture long-range dependencies between each item \cite{vaswani2017SA}. Several recent studies adopt self-attention techniques in many computer vision tasks. For example, Wang \etal \cite{wang2018nonlocal} utilize non-local operations to capture long-term information for the video classification task.
Besides, self-attention is utilized in object detection and image classification tasks \cite{carion2020detr, dosovitskiy2020, zhao2020san}, and prove that the effectiveness of capturing long-range dependency. Additionally, the self-attention operation can also be integrated with the generative model for image generation tasks \cite{Zhang2018SelfAttentionGA}. 

\section{Proposed Method}
\subsection{Problem Definition}
Given two monocular images between two timestamps $t-1$ and $t$ with known camera parameters, we aim to estimate the vehicle's distance and velocity in the current frame $t$ related to the camera coordinate. For each vehicle, the inter-vehicle distance $d \in \mathbb{R}^+$ is defined as the distance from the onboard camera optical center to the closest tangent plane on the vehicle surface, which is shown in Figure \ref{fig:arch} (top right). We take the planar position of the closest point on the vehicle in meter,  $p \in \mathbb{R}^3$, as the vehicle position in the current frame $t$. Assuming the corresponding closest points on the vehicle in the last frame $t-1$ is $p' \in \mathbb{R}^3$, the relative velocity is defined as $v = (p - p') / \Delta{t} \in \mathbb{R}^3$, where $\Delta t$ is the time difference between last frame and current frame in seconds. Our goal is to estimate the vehicle's state $\zeta=(v,p)$ including its position and velocity in current frame $t$.

\subsection{Overview}
\begin{figure*}[t]
\begin{center}
\includegraphics[scale=0.47]{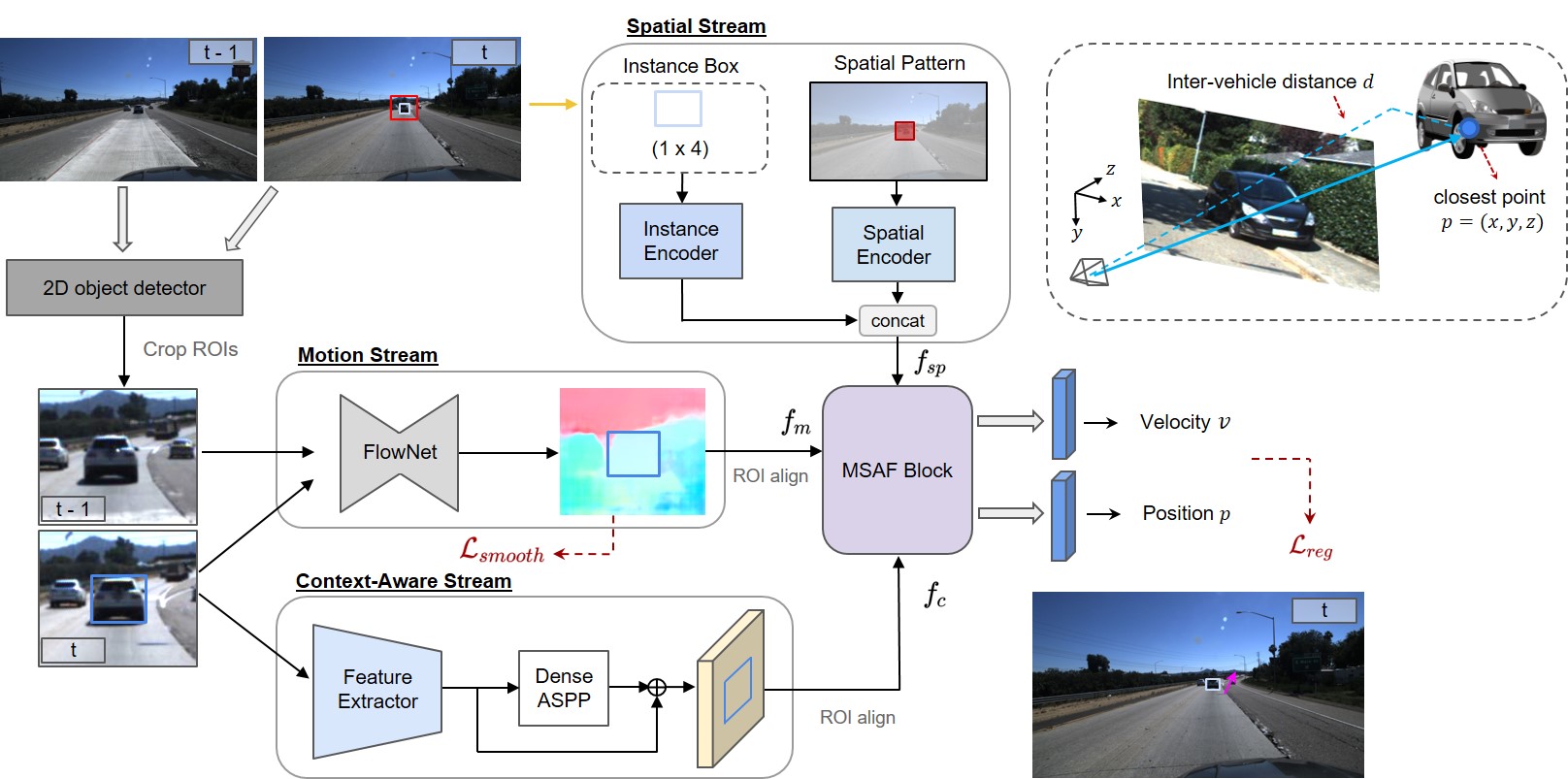}
\end{center}
   \caption{The architecture of MSANet. We adopt the motion stream, visual stream, and spatial stream to represent different views of the vehicle (Section \ref{sec:represent}). A multiple streams attention fusion (MSAF) block is proposed to merge different features to benefit the vehicle's state estimation task (Section \ref{sec:fusion}).}
\vspace{-0.7cm}
\label{fig:arch}
\end{figure*}

The overall architecture of our MSANet for joint vehicle velocity and position estimation is shown in Figure \ref{fig:arch}. Given two monocular images from two timestamps, we first detect the vehicles by an off-the-shelf object detector (\eg, Faster RCNN \cite{ren15fasterrcnn}), followed by cropping the expanded region of the vehicle. Next, we apply three kinds of encoders to extract different information, including motion clues, context-aware features, and spatial patterns (Section \ref{sec:represent}). Then, a multiple streams attention fusion (MSAF) block is proposed to fuse all features effectively (Section \ref{sec:fusion}). Finally, the intermediate fused representation from the MSAF block can be used to predict the position and velocity of the vehicle. Besides, a global relative constraint (GLC) loss is further proposed to encourage learning consistency between vehicles (Section \ref{sec:loss}).  

\subsection{Multiple Streams Feature Representation} \label{sec:represent}
\noindent{\bf Motion stream.} Similar to prior works \cite{Kampelmuhler2018camera,song2020endtoend}, to estimate the vehicle motion, we use an optical flow network to learn the dense motion clues of the cropped vehicle from the previous frame $I_{t'}$ to the current frame $I_t$. Following the object detection methods, ROI Align \cite{he2017maskrcnn} technique is applied to extract the object features inside the vehicle region. 
The motion clue $f_m$ can be formulated as:
\begin{equation}
\vspace{-0.05cm}
f_m = RoI({\bf F_m}, x_v)
\label{eq:1}
\vspace{-0.1cm}
\end{equation}
where $\bf F_m$ is the optical flow feature, and $x_v$ is the area of the vehicle.

\smallskip\noindent{\bf Context-Aware stream.}
We apply a backbone CNN as an encoder to extract the visual information. To further refine the features, we applied a DenseASPP \cite{yang2018denseaspp}, which combines dense connection skills \cite{huang2017densely} to ASPP \cite{chen2017deeplab} to increase larger receptive field. It can benefit to enhance the network. We also integrate the residual module \cite{He2016deep} with DenseASPP to make training stable and improve performance. Finally, the ROI align technique is adopted to obtain the context-aware feature $f_c$:
\begin{equation}
\vspace{-0.1cm}
f_c = RoI({\bf\mathcal{D}(F_c) + F_c}, x_v)
\label{eq:2}
\vspace{-0.05cm}
\end{equation}
where $\bf F_c$ is the appearance feature from CNN block, and $\mathcal{D}$ is the DenseASPP \cite{yang2018denseaspp} module.

\smallskip\noindent{\bf Spatial stream.} 
Intuitively, the size and location of the bounding box benefit to predict the position and velocity of the vehicle. We first encode the 2D bounding box with box center $(b_x, b_y)$ and its width $b_w$ and height $b_h$. With the known camera focal length $(f_x, f_y)$ and the principle point $(c_x, c_y)$, the box can be transformed from the pixel coordinate to the world coordinate ${\bf p} = [p_x, p_y, p_w, p_h]$ as follows:
\begin{equation}
\vspace{-0.05cm}
p_x = \frac{(b_x-c_x)}{f_x}\hat{z}, p_y = \frac{(b_y-c_y)}{f_y}\hat{z}, p_w = \frac{b_w}{f_x}, p_h = \frac{b_h}{f_y}
\label{eq:2}
\vspace{-0.1cm}
\end{equation}
where $\hat{z}$ is a fixed depth scalar. Next, we apply an instance encoder with a two-layer fully connected layer to encode the instance box feature $f_i$.

On the other hand, the spatial pattern of the vehicle is also an important clue: The bounding box position of a farther vehicle is typically closer to the top in the image. To this end, we design an encoder to extract the spatial pattern. Given a vehicle bounding box region, we generate a one-channel binary map representing the front object and background. The map has zeros everywhere except the vehicle location. We apply a spatial encoder with two layers of convolution kernels, followed by a global average pooling operation and a linear transform to obtain the spatial pattern feature $f_p$.

Finally, the final spatial features $f_{sp}$ are generated by simply concatenate instance feature $f_i$ and spatial pattern feature $f_p$:
\begin{equation}
\vspace{-0.05cm}
f_{sp} = f_i \parallel f_p
\label{eq:2}
\vspace{-0.05cm}
\end{equation}
where $\parallel$ means the concatenate operation along the channel axis.

\subsection{Multiple Streams Attention Fusion}\label{sec:fusion}
\begin{figure*}[t]
\begin{center}
\includegraphics[scale=0.47]{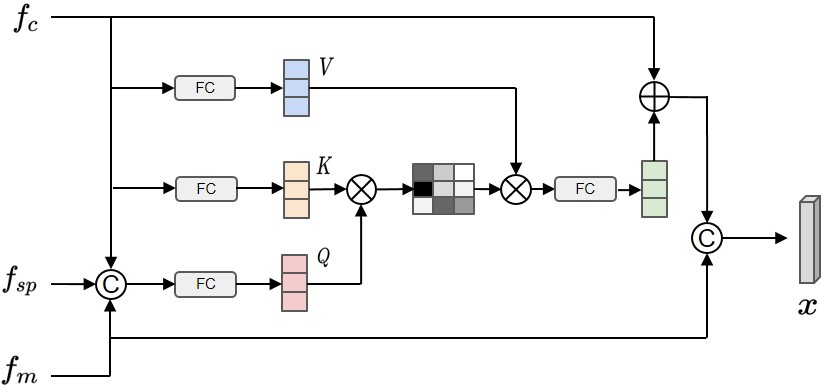}
\end{center}
   \caption{Multiple streams attention fusion block. The block fuses context feature $f_c$, spatial feature $f_{sp}$, and motion feature $f_m$ to generate the final representation feature $x$.}
\vspace{-0.85cm}
\label{fig:anl}
\end{figure*}
Inspired by \cite{wang2019atloc} that integrates the self-attention module to refine the flattened input feature, we share the same flavor and extend it to build a multi-stream attention fusion (MSAF) block to fuse different features, which is shown in Figure \ref{fig:anl}. Three flattened features are concatenated firstly, and the hybrid feature $Q$ is generated by a linear transform $W_{Q}$:
\begin{equation}
\quad Q = W_{Q}(f_c \parallel f_m \parallel f_{sp})
\label{eq:3}
\end{equation}
We attempt to calculate the correlation between hybrid feature $Q$ and context-aware feature $f_c$, which is different from the standard non-local module. Simply transform context features with two non-shared FC layers $W_K$ and $W_V$, two intermediate features $K$ and $V$ can be obtained. Followed by combining with hybrid feature $Q$ to generate the attention map $S$ by matrix multiplication and also the attentive output $F$ as follows:
\begin{equation}
\vspace{-0.1 cm}
\begin{gathered}
S = softmax(Q^TK) \\
F = SV
\vspace{-0.1 cm}
\end{gathered}
\end{equation}
Finally, the correlation feature $F$ is passed through a fully connected layer $W_F$ and added with a shortcut connection of the original contextual feature. Then, the above feature can concatenate with the motion feature to obtain the final squeezed feature $x$:
\begin{equation}
x = (f_{sp}+ W_F \cdot F) \parallel f_m
\label{eq:4}
\vspace{-0.1 cm}
\end{equation}
In this way, the final represented feature $x$ becomes more robust and is beneficial for predicting the vehicle's position and velocity (Section \ref{sec:ablation}).

\subsection{Global Relative Constraint Loss}\label{sec:loss}

\begin{figure*}[!ht]
\vspace{-0.3 cm}
\begin{center}
\includegraphics[scale=0.505]{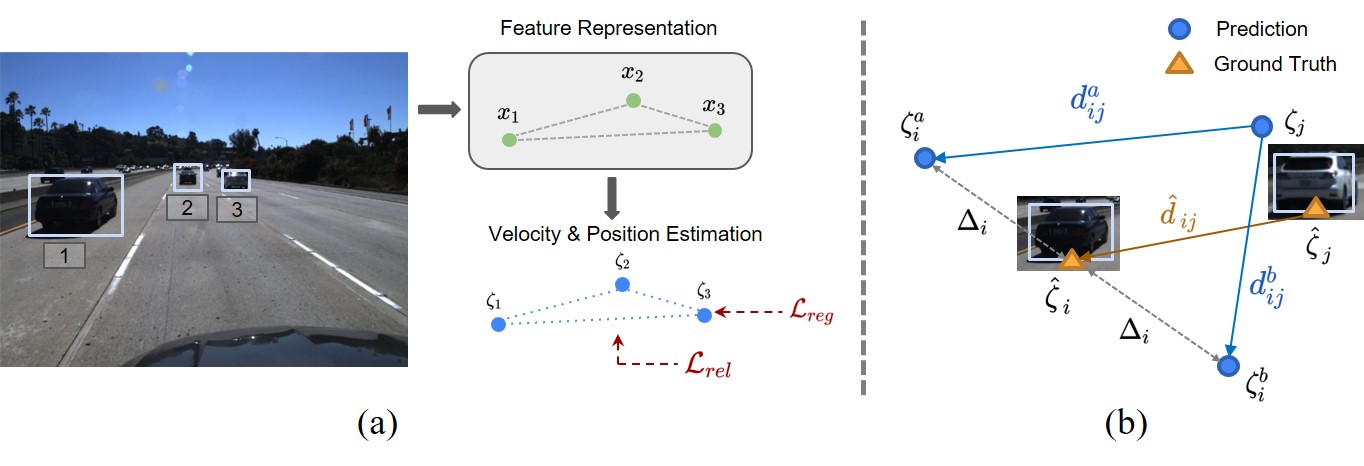}
\vspace{-1 cm}

\end{center}
   \caption{Vehicle regression example. (a) We regress each vehicle's velocity and position by the regression loss ($L_{reg}$) and incorporate relative global constraint (GLC) loss ($L_{rel}$) to regularize the predictions between vehicles. (b) Two predictions $\zeta^a_{i}$ and $\zeta^b_{i}$ contribute the same regression loss to the vehicle $i$ but have different relative state $d^a_{ij}$ and $d^b_{ij}$ to the vehicle $j$ with predicted state $\zeta_{j}$, which generate different relative loss for true relative state $\hat d_{ij}$.
   }
\label{fig:rel_loss}
\end{figure*}

As shown in Figure \ref{fig:rel_loss}(a), our model aims at regressing each vehicle's velocity and position by a regression loss $L_{reg}$, which is similar to \cite{song2020endtoend}. 
Due to predicting the absolute state of the vehicle from the monocular camera is difficult, we further incorporate relative global constraint (GLC) between vehicles to enforce prediction consistency, which helps the model reduce errors by imposing constraints of relative vehicle states.

An illustration of relative global constraint (GLC) is shown in Figure \ref{fig:rel_loss}(b). 
We consider two different predictions $\zeta^a_{i}$ and $ \zeta^b_{i}$ with the same shift $\Delta_{i}$ to the vehicle true state $\hat\zeta_{i}$, they contribute the same regression loss for vehicle $i$.
However, for another vehicle $j$ with predicted state $\zeta_{j}$, we observe that the prediction $\zeta^a_{i}$ is more reasonable than $\zeta^b_{i}$ because the state difference $d^a_{ij}$ is more close to true state difference $\hat d_{ij}$ than the other one $d^b_{ij}$. To this end, we propose a relative loss to explore the state difference between vehicles, which helps the model enforce global consistency and improve velocity and position prediction performance.
The {\bf global relative constraint (GLC) loss} is formulated as:
\begin{equation}
\vspace{-0.2 cm}
L_{rel}=\sum_{i,j = 1,i \neq j}h(d_{ij},\hat d_{ij})
\label{eq:res}
\end{equation}
\vspace{-0.2 cm}
where $d_{ij} = (\zeta^{i}_{v}-\zeta^{j}_{v}, \zeta^{i}_{p}-\zeta^{j}_{p})$ is the relative state between vehicles. $h(\cdot)$ is noted as the function to measure the distance. $\hat{(\cdot)}$ represents the ground truth of the relative state. 
For distance function $h(\cdot)$ choice, we choose the Charbonnier loss \cite{barron2019general} as the objective function, $L_{Cha}$, which is a robust $L_1$ loss, can be expressed as follows:
\begin{equation}
\vspace{-0.1cm}
h(\bf{s}, \bf{\hat s}) = \sqrt[]{({\bf s}- {\bf{\hat s}})^2+\epsilon ^2}
\label{eq:res}
\vspace{-0.1cm}
\end{equation}
where $\epsilon$ is a small constant (\eg,$10^{-6}$). The Charbonnier loss function can handle the outliers efficiently, which benefits network training (Experimental results in Section \ref{sec:ablation}). 

To train the MSANet, we apply a regression loss function to estimate the absolute velocity and position of each vehicle:
\vspace{-0.3cm}
\begin{equation}
\vspace{-0.2cm}
L_{reg}=\sum_{i}^{N}h(\zeta^{i}_{v},\hat \zeta^{i}_{v})+\lambda\sum_{i}^{N}h(\zeta^{i}_{p},\hat \zeta^{i}_{p})
\label{eq:pv}
\vspace{-0.2cm}
\end{equation}
where $\lambda$ is the scaling coefficient between position and velocity, which is set as 0.1 as default. $N$ is the number of target vehicles in the image. Furthermore, similar to \cite{Godard2017monodepth}, to encourage smoother optical flow prediction, we adopt a {\bf smoothness loss} for estimated optical flow $F$ with the cropped image $I$:
\vspace{-0.2cm}
\begin{equation}
L_{smooth}=\sum_{i,j}\sum_{d \in x, y}|\partial_{d}F(i,j)|e^{-|{\partial_{d}}I(i,j)|}
\label{eq:smooth}
\vspace{-0.2cm}
\end{equation}
where $d$ represents partial derivative on $x$ and $y$ direction. This loss is an edge-aware loss, which can help to enforce local smoothness for optical weighted by image gradients.

The final loss function can be summarized as a weighted sum of the above four terms:
\vspace{-0.2cm}
\begin{equation}
L = L_{reg} + \lambda_{1}L_{smooth} + \lambda_{2}L_{rel}
\label{eq:res}
\vspace{-0.2cm}
\end{equation}
where $\lambda_1$ and $\lambda_2$ are the scaling coefficients. We set $\lambda_1 = 1$, $\lambda_2 = 0.3$ as default.
\section{Experiments}

\subsection{Experimental Setup}

\noindent{\bf Dataset.}
We adopt two datasets for experiments: KITTI raw dataset \cite{Geiger12CVPR} and Tusimple velocity dataset $\footnote{https://github.com/TuSimple/tusimple-benchmark}$. Tusimple velocity dataset includes 1074 driving sequences for training, of which the video length is 2 seconds under 20 fps. The bounding boxes are annotated for the last frame. For the KITTI raw dataset, we follow the setting in \cite{song2020endtoend}. Due to the detailed tracklet of each vehicle in the video clips are available, we can generate distance and velocity ground truth by ourselves.

\smallskip\noindent{\bf Training details.}
MSANet utilizes PWC-Net \cite{Sun2018PWC-Net} pretrained on the FlyingChairs \cite{Dosovitskiy2015Flownet} as the optical flow extractor to estimate the motion information of the vehicle and adopts ResNet34 \cite{He2016deep} as our feature extractor.  During training, We utilize the Adam optimization algorithm with mini-batch 4. The total epoch is 100. Whole experiments are implemented in Pytorch.

\smallskip\noindent{\bf Metrics.}
For velocity estimation, we follow the rules in the Tusimple Velocity Challenge. The vehicle velocity are categorized into three groups according to their relative distance $d$ to the ego-vehicle: near-range ($d < 20 m$), medium-range ($20 m < d < 45 m$), and far-range ($d > 45 m$). 
The main metric is the Mean Square Error (MSE) of the velocity with the unit $m^2/s^2$ and position with the unit $m^2$, following the metric used in the Tusimple dataset. The mean MSE of the three groups is regarded as the final metric.
For vehicle distance estimation, we utilize the standard metrics in the depth estimation tasks \cite{Fu2018dorn, Eigen2014depth}, including absolute relative difference (AbsRel), square relative difference (SqRel), RMSE, RMSE (log), and $\delta$. It is noted that we only consider the closet point for each vehicle, which follows the setting described in \cite{song2020endtoend}.

\subsection{Main Results}
\begin{table}[t]
\footnotesize
\centering
\begin{tabular}{c|c|cccc}
\hline
                         & \multicolumn{1}{c|}{Position} & \multicolumn{4}{c}{Velocity} \\ \cline{2-6} 
                         	 & MSE $\downarrow$	  & MSE (near)$\downarrow$   & MSE (medium)$\downarrow$	  & MSE (far)$\downarrow$   & MSE (avg.)$\downarrow$	 \\ \hline \hline
Rank1 \cite{Kampelmuhler2018camera}   & - & 0.18 & 0.66 & 3.07 & 1.30 \\
Rank2   & - & 0.25 & 0.75 & 3.50 & 1.50 \\
Rank3   & - & 0.55 & 2.21 & 5.94 & 2.90 \\ 
Song \etal ($org$) \cite{song2020endtoend}   & 9.72 & 0.23 & 0.99 & 3.27 & 1.50  \\
Song \etal ($full$) \cite{song2020endtoend}  & 10.23 & 0.15 & 0.34 & 2.09  & 0.86  \\  \hline
\bf{Ours}  & \bf{7.56} & \bf{0.10}  & \bf{0.26}   & \bf{1.58}  & \bf{0.65}  \\  \hline
\end{tabular}
\vspace{-0.45cm}
\caption{{The quantitative results of vehicle position and relative velocity estimation on the Tusimple dataset.}}
\label{Tab:tus_vel}
\end{table}

\setlength{\tabcolsep}{0.011\linewidth}{
\begin{table*}[t]
\vspace{-0.2cm}
\footnotesize
\centering
\begin{tabular}{c|ccccccc} \hline
                     &AbsRel$\downarrow$&SqRel$\downarrow$&RMSE$\downarrow$&RMSE (log)$\downarrow$&$ \delta<1.25^1\uparrow$ & $\delta<1.25^2\uparrow$ & $\delta<1.25^3\uparrow$	 \\ \hline \hline
Song \etal ($org$) \cite{song2020endtoend}    & 0.037 & 0.132 & 2.700 & 0.059 & 0.989 & \bf{1.000} & \bf{1.000} \\
Song \etal ($full$) \cite{song2020endtoend}   & 0.041 & 0.152 & 2.894  & 0.062 & 0.987 & \bf{1.000} & \bf{1.000} \\  \hline
\bf{Ours}  & \bf{0.034}  & \bf{0.105}   & \bf{2.416}  & \bf{0.050} & \bf{0.997} & \bf{1.000} & \bf{1.000}\\  \hline
\end{tabular}
\vspace{-0.45cm}
\caption{{The quantitative results of vehicle distance estimation on the Tusimple dataset.}}
\label{Tab:tsu_pos}
\end{table*}
}

\noindent{\bf Experimental results on Tusimple dataset.}
The velocity and position in the Tusimple dataset are annotated along two directions x-axis and z-axis of the camera coordinate. Following the setting in \cite{song2020endtoend}, we set the output dimension of the proposed network as three dimensions, including one for position and two for velocity. The remaining dimension of position can be obtained by inverse projection with the bounding box of the vehicle. 
In practice, we transform the center bottom pixel of the cropped image to the world coordinate to obtain the z-axis reference position and learn the residual value from the network to predict the vehicle’s z-axis position.

To show the capability of our proposed network, we compare to the top 3 ranked in Tsusimple velocity challenge and a joint velocity and position estimation network proposed in \cite{song2020endtoend}. As shown in Table \ref{Tab:tus_vel}, our model has superior performance than others. The rank one method \cite{Kampelmuhler2018camera} in the challenge utilizes three separate models for different ranges, which under the risk for hyperparameter tuning.
In \cite{song2020endtoend}, the authors predict all range distance and velocity from the cropped vehicle image, which will cause the performance limit.
On the contrary, we additionally leverage the spatial and context-aware information and predict the vehicle's state jointly with the global relative constraint (GLC) loss, which benefits the model to predict the different ranges.
The average mean square velocity error of our approach is about 0.65 $m^2/s^2$ corresponded to 0.40 $m/s$ absolute error, which is better than the method in \cite{song2020endtoend} (0.86 $m^2/s^2$ MSE corresponded to 0.48 $m/s$ absolute error).
Besides, the detailed statistic of the vehicle distance regression performance is shown in Table \ref{Tab:tsu_pos}. 
It proves that the proposed network and global relative constraint (GLC) are effective for distance estimation.

\smallskip\noindent{\bf Experimental results on KITTI dataset.}
In Table \ref{Tab:kitti_vel}, we present our model's performance of velocity prediction on the KITTI dataset. Our method outperforms prior art \cite{song2020endtoend} among all distance ranges for relative velocity estimation.
We further report the results of distance estimation on the KITTI dataset with the same setting in \cite{song2020endtoend} for a fair comparison. As shown in Table \ref{Tab:kitti_pos}, our approach gets the competitive results and outperforms the others in most metrics. Moreover, our model gets less outlier due to the model predicting the reasonable estimation with the global relative constraint (GLC).

\begin{table*}[h]
\footnotesize
\centering
\begin{tabular}{c|cccc} \hline
                         	 & MSE (near)$\downarrow$   & MSE (medium)$\downarrow$	  & MSE (far)$\downarrow$   & MSE (avg.)$\downarrow$	 \\ \hline \hline
Song \etal ($full$) \cite{song2020endtoend}   & 0.29 & 0.93 & 1.57  & 0.94  \\  \hline
\bf{Ours}  & \bf{0.23}  & \bf{0.67}   & \bf{0.96}  & \bf{0.62}  \\  \hline
\end{tabular}
\vspace{-0.45cm}
\caption{{The quantitative results of velocity estimation on KITTI dataset.}}
\label{Tab:kitti_vel}
\end{table*}
\setlength{\tabcolsep}{0.013\linewidth}{
\begin{table*}[ht]
\vspace{-0.5cm}
\footnotesize
\centering
\begin{tabular}{c|ccccccc} \hline
                     &AbsRel$\downarrow$&SqRel$\downarrow$&RMSE$\downarrow$&RMSE (log) $\downarrow$& $ \delta<1.25^1\uparrow$ & $\delta<1.25^2\uparrow$ & $\delta<1.25^3\uparrow$	 \\ \hline \hline
3DBBox \cite{Mousavian3Dbbox}    & 0.222 & 1.863 & 7.696 & 0.228 & 0.659 & 0.966 & 0.994 \\
DORN \cite{Fu2018dorn}    & 0.078 & 0.505 & \bf{4.078} & 0.179 & 0.927 & 0.985 & 0.995 \\
Unsfm \cite{bian2019depth}    & 0.219 & 1.924 & 7.873 & 0.338 & 0.710 & 0.886 & 0.933 \\
Song \etal \cite{song2020endtoend}   & \bf{0.075} & 0.474 & 4.639  & \bf{0.124} & 0.912 & 0.996 & \bf{1.000}  \\  \hline
\bf{Ours}  & 0.098  & \bf{0.444}   & 4.240  & 0.127  & \bf{0.930} & \bf{0.998} & \bf{1.000} \\  \hline
\end{tabular}
\vspace{-0.45cm}
\caption{{The quantitative results of vehicle distance estimation on the KITTI dataset. We compare with baseline results of prior works reported in \cite{song2020endtoend}.}}
\label{Tab:kitti_pos}
\end{table*}
}

\smallskip\noindent{\bf Qualitative result.}
Figure \ref{fig:visual} gives the visualization of prediction results for position and vehicle on the Tusimple dataset. The example shows that the prediction of our model is closed to the ground truth. Moreover, it also has less error compared to \cite{song2020endtoend}, which indicates the proposed methods have better performance for jointly predicting velocity and position. 
\setlength{\tabcolsep}{0.009\linewidth}{
\begin{figure}[!htb]
\scriptsize
\centering
\begin{minipage}[h]{\textwidth}
\centering
\includegraphics[scale = 0.47]{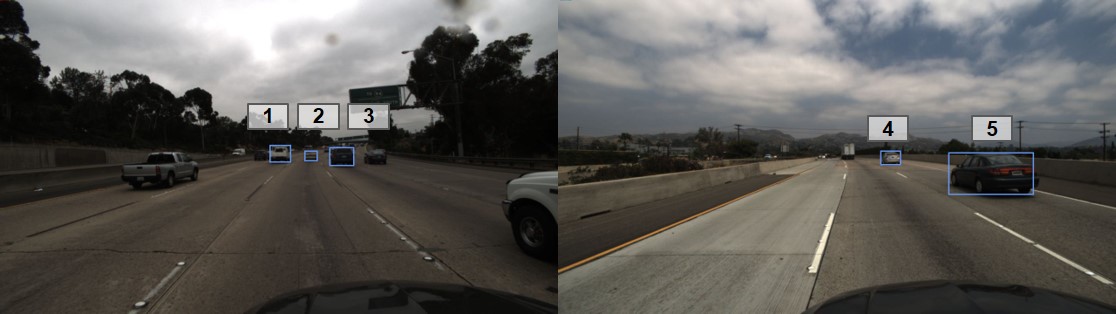}
\caption{Qualitative visualization results on Tusimple dataset. MSE scores are shown as $mean\pm \sqrt {variance}$.}

\label{fig:visual}
\end{minipage}\hfill
\begin{minipage}[t]{0.9\textwidth}
\centering
\vspace{3pt}
\scriptsize
\begin{tabular}{c|ccc|ccc}
\hline
                         & \multicolumn{3}{c|}{Position (m)} & \multicolumn{3}{c}{Velocity (m/s)} \\ \cline{2-7} 
                     Vehicle id   & \cite{song2020endtoend} & Ours  & GT & \cite{song2020endtoend} & Ours  & GT	 \\ \hline \hline
1  & (35.1, -1.6) & (33.9, -1.5) & (36.9, -1.8) & (4.3, -0.5) & (4.7, -0.3) & (5.5, -0.2) \\
2  & (44.2, 2.3) & (42.6, 2.3) & (41.2, 0.7) & (1.3, -0.2) & (1.8, -0.1) & (2.1, -0.1) \\
3  & (30.5, 4.6) & (26.9, 4.1) & (27.3, 2.7) & (0.2, -0.5) & (-0.0, -0.3) & (0.1, -0.4) \\ \hline
4  & (36.5, 4.5) & (34.2, 4.2) & (34.0, 2.7) & (-0.2, -0.0) & (-0.7, -0.4) & (-0.5, -0.2) \\
5  & (11.9, 5.3) & (10.6, 4.7) & (9.8, 3.3) & (-1.2, -0.3) & (-1.0, -0.1) & (-0.7, -0.3) \\

\hline\hline
MSE $\downarrow$  & 9.23 $\pm$ 3.71 ($m^2$) & \bf{4.10 $\pm$ 2.64 ($m^2$)} & - & 0.48 $\pm$ 0.54 ($m^2/s^2$) & {\bf 0.16 $\pm$  0.18 ($m^2/s^2$)} & -  
\vspace{-0.45 cm}
\end{tabular}
\end{minipage}
\end{figure}
}

\subsection{Ablation Study} \label{sec:ablation}
\noindent{\bf Effect of Proposed Streams and Fusion Module.}
We conduct an ablation study to analyze the effect of the proposed three streams and fusion block. The experimental results are shown in Table \ref{tab:abl_stream}. Firstly, only the motion stream (M) is left, and it got an undesirable performance, which achieves 0.91 (MSE) for velocity estimation. Secondly, the spatial stream (SP) is further adding to the network. 
The MSE becomes 0.84, which shows that extra spatial information helps to improve model performance. Thirdly, we combine context-aware stream (C) to the network by simply concatenating operation. The performance improved by 10\% achieving 0.75, demonstrating that the contextual information is helpful for velocity and position estimation. Finally, we fuse three streams with the proposed multiple-stream attention fusion block, and the entire network achieves the best performance at 0.65. The experimental results prove the effectiveness of the three streams and the fusion block.

\smallskip\noindent{\bf Effect of Proposed Losses.}
To investigate the efficiency of the proposed loss, we conduct five experiments with different losses, which are listed in Table \ref{tab:abl_loss}. The first three rows compared different distance measurement function $h(\cdot)$ described in Eq. \ref{eq:pv}. As we can see, we prove the claim that using Chabonnier loss will have a better performance compared to L1 and Smooth L1 loss. The fourth row shows the impact of smoothness loss. Furthermore, the last row shows that the proposed relative constraint loss helps to regularize the consistency between the vehicles in the same frame, which achieves better performance. 

\setlength{\tabcolsep}{0.012\linewidth}{
\begin{table*}[h]
\footnotesize
\RawFloats
\centering
\makebox[0pt][c]{\parbox{1\textwidth}{%
    \begin{minipage}[b]{0.485\hsize}\centering
    \begin{tabular}{c|ccc|c|c}
    \hline
    Index & M & SP & C & MSE (V) $\downarrow$ & MSE (P) $\downarrow$ \\ \hline \hline
    1  & $\surd$  &\large$\times$& \large$\times$ & 0.91 & 10.23 \\ \hline
    2  & $\surd$   &  $\surd$  &\large$\times$ & 0.84  & 8.26  \\ \hline
    3  &  $\surd$   &  $\surd$    &$\surd$  & 0.75   & 8.02    \\ \hline
    \underline{4}  &  $\surd$   &  $\surd$    &$\surd$ &  {\bf 0.65}  &   {\bf7.56}   \\ \hline

    \end{tabular}
    \caption{The ablation study shows effeteness of motion stream (M), spatial stream (SP), and context stream (C), and also the proposed fusion block. The underline indicates using the proposed fusion block.}
    \label{tab:abl_stream}
       
    \end{minipage}
    \hfil
    \begin{minipage}[b]{0.49\hsize}\centering
\begin{tabular}{l|c}
\hline
Loss Function       & MSE (V) $\downarrow$  \\ \hline\hline
$L_{reg} (L_1)$       & 0.85\\
$L_{reg} ($Smooth $L_1)$       & 0.79  \\
$L_{reg} (L_{Cha})$     &  0.77 \\
$L_{reg} (L_{Cha}) + L_{Smooth}$     & 0.73   \\
$L_{reg} (L_{Cha}) + L_{Smooth} + L_{Rel}$         &  {\bf 0.65}    \\ \hline
\end{tabular}
        \vspace{+0.15cm}
        \caption{The analysis on different loss functions of the proposed framework. $L_{reg}(\cdot)$ means the different measurement function $h$ described in Eq. \ref{eq:pv}.}
        \label{tab:abl_loss}

    \end{minipage}
    \hfil
}}
\end{table*}
}

\section{Conclusion}
This work presents a novel framework for joint vehicle velocity and inter-vehicle distance estimation. MSANet leverages multiple information, including context-aware features, motion clues, and spatial positions to learn the vehicle's state. A novel global relative constraint (GLC) loss is proposed to resolve the prediction inconsistency problem. 
Experiments on KITTI and Tusimple datasets validate the effectiveness of the proposed approach.
We believe our idea paves a new path for research of advanced driver-assistance systems.

\section*{Acknowledgements}
This work was supported in part by the Ministry of Science and Technology, Taiwan, under Grant MOST 110-2634-F-002-026. We are grateful to the National Center for High-performance Computing. We also thank Tsung-Han Wu for his helpful discussions on this work.

\bibliography{egbib}
\end{document}